\documentclass[aps,twocolumn,amsmath,amssymb,nofootinbib,superscriptaddress,floatfix]{revtex4-1}
\usepackage{amssymb}
\usepackage{mathrsfs}
\usepackage{color}
\usepackage{graphicx}
\usepackage{srcltx}
\usepackage{soul}
\usepackage{makecell}
\usepackage[hidelinks]{hyperref}

\begin{document}
\title{Knowledge-based Document Classification with Shannon Entropy}
\author{AtMa P.O. Chan}
\affiliation{Microsoft, Redmond, Washington, 98052, USA}

\begin{abstract}
Document classification is the detection specific content of interest in text documents. In contrast to the data-driven machine learning classifiers, knowledge-based classifiers can be constructed based on domain specific knowledge, which usually takes the form of a collection of subject related keywords. While typical knowledge-based classifiers compute a prediction score based on the keyword abundance, it generally suffers from noisy detections due to the lack of guiding principle in gauging the keyword matches. In this paper, we propose a novel knowledge-based model equipped with Shannon Entropy, which measures the richness of information and favors uniform and diverse keyword matches. Without invoking any positive sample, such method provides a simple and explainable solution for document classification. We show that the Shannon Entropy significantly improves the recall at fixed level of false positive rate. Also, we show that the model is more robust against change of data distribution at inference while compared with traditional machine learning, particularly when the positive training samples are very limited. 
\end{abstract}

\maketitle

\textbf{1.$~$Introduction}---Document classification is the classification of text content in documents. It detects the domain specific concepts and suggests the underlying content. Such content classification task has long been a famous problem in computer science and information retrieval. As the data volume in cloud storages has been growing at an exponential rate in the recent decades, such classification problem is becoming more and more important in enterprise data management aspects, e.g. information governance, sensitive information protection, data loss prevention, and ediscovery. Document classifiers automate the concept extraction process, categorizing vast amount of dark data so that organizational policies can be applied to the appropriate documents.

Supervised machines learning model has been very promising for documents classifications given the availability of large amount of labeled data. For example, one can train a classifier that classify news content based on the categorized news corpus. However, this method is limited to categories that is generally available, e.g. finance, sport, politics, entertainment etc. Labeled data for more sensitive content is very limited, where supervised models tend to over-fit at training and the performance usually breaks down at inference. Large language models \cite{devlin-etal-2019-bert, Liu2019RoBERTaAR} have been proposed to train classifiers based on small amount of labeled data with transfer learning. However, these pretrained models, even with distillation \cite{10.1145/1150402.1150464,Hinton2015DistillingTK}, have rather large compute usage, model size and latency, making them not scalable to large data.

\begin{figure}
\centering
\includegraphics[scale=0.3]{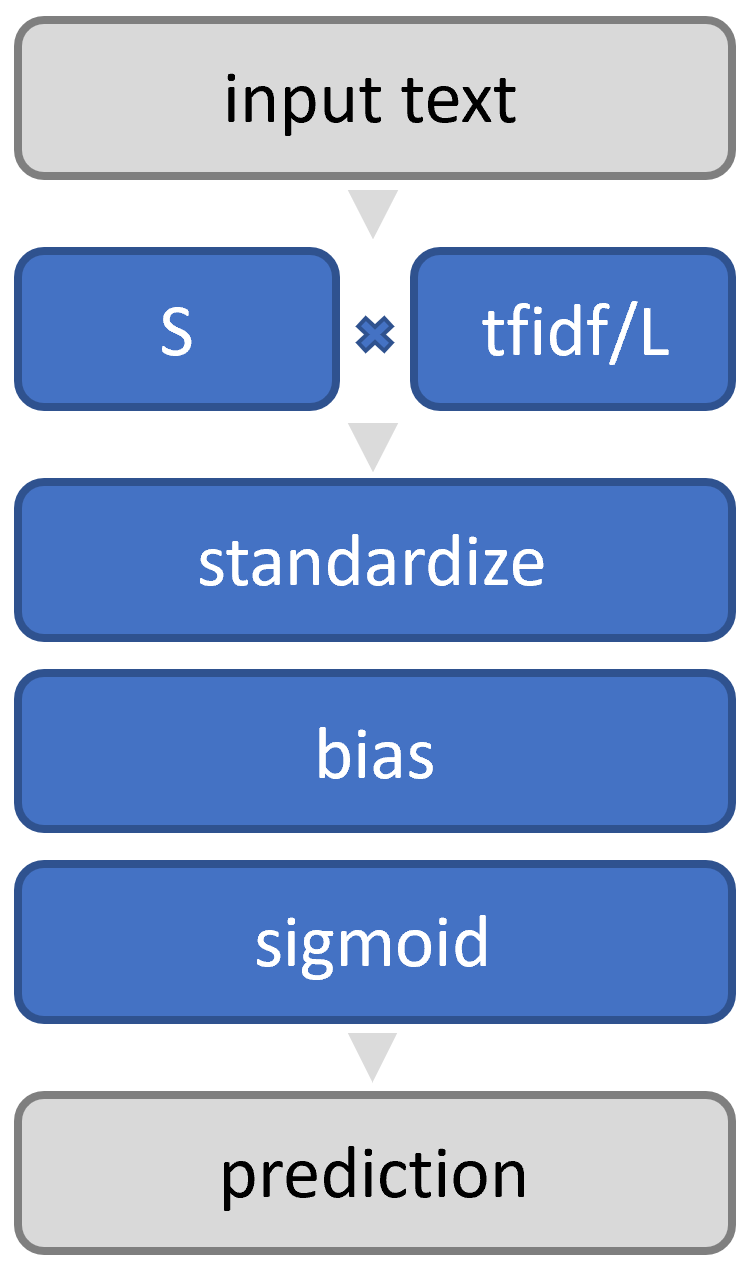}
\caption{Architecture of the knowledge-based model with Shannon Entropy. The model is associated with a collection $G$ of keywords from a target category. Given the text $x$ of an input document, the model first measures the keyword abundance from $G$ in the text $x$ by computing the $tfidf$ per document length $L$. Then it measures the richness of information by computing the Entropy $\mathcal{S}$. The multiplication of the two quantities gives the score of the text. The score is then standardized and offset by a bias $b$. A sigmoid is operated on the final score to make the prediction a probability [Eq. (\ref{prediction})].}\label{Fig1}
\end{figure}

Alongside with the data-driven machine learning classifiers, knowledge-based model has been proposed to classifies documents without invoking labeled data \cite{10.1007/978-3-030-21348-0_4,inproceedings,10.3115/1118824.1118844,miller-etal-2016-unsupervised,haj-yahia-etal-2019-towards,Trker2018TECNEKB,10.1007/978-3-030-21348-0_23}. Typically, such model intakes knowledge in the form of a collection of subject specific keywords. For any document, the algorithm computes a score based on the abundance of keyword matches in the document. While this method is simple and scalable, the detection is usually noisy due to the lack of guiding principle in keyword detection, e.g. a document can be classified as with certain content because of a large count of a single common keyword. Such issue arises because the abundance measure does not take care of information richness or diversity.

Meanwhile, Entropy is well-known to be a fundamental measure of information content. It was first introduced in thermodynamics and statistical mechanics to characterize the randomness of physical systems. It was brought to information science to formulate a mathematical theory for communication \cite{6773024,6773067}. Such information Entropy, known as Shannon Entropy, measures the information richness of a given distribution, quantifying the uncertainty or the level of surprise. In classical information theory, Entropy measures the information content within a message, establishing a theoretical limit to how much a message can be losslessly compressed \cite{10.5555/971143,10.5555/1146355}. In quantum information theory, its quantum analogue, Von Neumann Entropy, measures the information content of a quantum system or the quantum entanglement information in a bipartite system.

In this paper, we propose a novel architecture for the knowledge-based model with Shannon Entropy [Fig. \ref{Fig1}], appropriately taking into account both the concept abundance and information richness.  Given a glossary of keywords from a target category, we construct a document classifier that classifies such target content. For any document, the model computes a prediction probability based on the concept abundance (tfidf/L) and concept diversity (Entropy). This method provides a simple and explainable solution for document classification without involving any positive target sample. We show that the recall of the knowledge-based model is significantly improved with the Entropy measure. Also, we show that the model is more robust against data variation while compared with traditional machine learning, particularly when the positive samples are rather limited. Here, we focus on binary document content classification without loss of generality. The remaining of this paper will be organized as follows. In section 2, we review related works on knowledge-based approach for categorizing documents. In section 3, we introduce the architecture of the knowledge-based classifier with Entropy. In section 4, we show the experimentation results. In section 5, we wrap up with a conclusion together with a couple of comments.

\textbf{2.$~$Related Work}---Here we review related works in knowledge-based approach for categorizing document content. Most of the previous knowledge-based methods stem from a collection of predefined domain specific keywords. Ref.\cite{10.1007/978-3-030-21348-0_4} combines exact match and semantic match approaches to retrieve top-K relevant documents given a knowledge graph of keywords. Ref.\cite{inproceedings} classifies document topic based on the document similarity with keywords automatically extracted from electronic magazines. Ref.\cite{10.3115/1118824.1118844} detects predefined entities in a text and assigns some small topics to each of the entities, it computes the final topic of the text by combining the small topics using an aggregation formula. Ref.\cite{miller-etal-2016-unsupervised} assign labels to sentences based on the occurrence of keywords and sentence similarity measure. Ref.\cite{haj-yahia-etal-2019-towards} describes an unsupervised approach for text categorization by calculating the similarity between a document and a category represented by a keyword list obtained by a series of enrichment process. Refs.\cite{Trker2018TECNEKB,10.1007/978-3-030-21348-0_23} embeds entities and the predefined categories in the same vector space using graph embedding. For a input text, the algorithm outputs a score for each category based similarity on the entities in the text and the categories. While most of the works measure the concept abundance based on work count or similarity measure, concept diversity is commonly ignored. To the best of our knowledge, it is the first of its kind to incorporate Entropy into knowledge-based document classification.

\textbf{3.$~$Knowledge-based Model}---Here, we present the architecture of the knowledge-based classifier. Given any target content category, the model intakes a glossary $G$ of domain specific keywords for the category, which can be easily obtained from subject matter experts or crawling from webs. Ideally, the glossary $G$ covers all concepts and topics in the target category. The model construction also involves an unsupervised training on a generic large background document corpus, which is not required to have any positive sample in the target category. For any input document $x$, the model calculates the abundance of keywords matches based on the normalized tfidf score,
\begin{align}
tfidf(x)/L=\sum_{w\in G}~tf_{w}(x)~idf_{w}/L~,
\end{align}
where $tf_{w}(x)$ is the term frequency of the keyword $w\in G$ inside the document $x$, and $idf_{w}$ is the inverse document frequency of the keyword $w$ computed in the background corpus, $L$ is the document length normalization. We use regularized length $L=$max($k$, \# of words in $x)$, which penalizes more on documents with number of words smaller than the parameter $k$. Based on the distribution $p_{w}$ of the keyword matches, the model calculates the information diversity by computing the Shannon Entropy,
\begin{align}
\mathcal{S}(x) = -\sum_{w\in G}~p_{w}\ln p_{w}~.
\end{align}
Note that the Entropy vanishes if and only if $p_{w}$ is a Dirac delta distribution. Hence the Entropy ignores keyword matches with only one keyword species. Also, given a fixed support for the distribution $p_{w}$, the Entropy is at maximum if and only if $p_{w}$ is uniform over the support. So the Entropy favors uniform keyword detections. In addition, given the distribution $p_{w}$ is uniform over a support, the larger the support is, the larger the Entropy. Hence the Entropy favors diverse keyword matches. The content score $s$ on the input $x$ is given by the product,
\begin{align}\label{score}
s(x) = \mathcal{S}(x)~tfidf(x)/L~.
\end{align}
Note that typical knowledge-based model consists only of the abundant measure. Here with the Entropy, the model naturally takes care of both concept abundance and information richness. In such construct, the content score $s$ is high if and only if keywords detected are abundant and diverse, making the model explainable in its detections. To compute the prediction score, we use the standardization $\hat{s}(x) = (s(x)-\mu)/\sigma$, where $\mu$ and $\sigma$ are respectively the mean and standard deviation of $s$ estimated in the background corpus. The final prediction probability is computed by an offsetting and a sigmoid,
\begin{align}\label{prediction}
y(x)=f\big(\hat{s}(x)-b\big)~,
\end{align}
where $b$ is the offsetting bias and $f$ is the sigmoid function that produces a probability $y\in[0,1]$. As a convention, $x$ is classified as having the target content if and only if $y(x)\ge0.5$, so the bias $b$ specifies the level of  $\hat{s}$ such that $x$ is classified as having significant target content. The parameter $b$ can be specified directly with regarded to its meaning, for example, $b=3$ means that significant content must have content score $s$ at least three standard deviation higher than mean wrt the background corpus. The value of $b$ can also be determined by requiring the model to achieve a certain false positive rate (FPR), which can be computed either on the large background dataset, or a separated faithful corpus if available. Remarkably, the whole construction of the model does not require the use of any positive target sample, which is generally difficult to get at for more sensitive categories.
 
\textbf{4.$~$Experiments}---In this section, we show the numerical results of the knowledge-based mode with the Shannon Entropy in two different experiments. In the first experiment, we show that the recall of the knowledge-based models is significantly improved with Entropy. In the second experiment, we show that the performance of such knowledge-based model is more robust against change of data distribution while compared with traditional data-driven machine learning approach.

In the first experiment, we consider knowledge-based models for nine target categories: business, tech, healthcare, legal, contract, hr, patent, tax, procurement. The glossary is crawled from web for each category, where the glossary size ranges from hundreds to thousands. We manually review each glossary to remove keywords with broad meaning. The background corpus here is composed of 100,000 randomly sampled articles from the Wikipedia dataset \cite{wikidump}. We take $k=100$ for the length normalization. Knowledge-based model is then constructed independently with and without Entropy. The bias $b$ for each model is tuned such that the FPR $=0.0005$ on a GPT-3 generated dataset \cite{brown2020language}, which composes of 4000 documents from 800 different categories. To evaluate the recall, 13,000 realistic document templates, with a variety of template types, are crawled for the nine categories.

Here, we study the impact of Entropy on the model performance. We measure the recall on the document templates for each category independently for knowledge-based model with and without Entropy [Table \ref{table1}]. From the results, we see the recall increased drastically with the Entropy measure over different categories, where the average recall increases from $0.216$ to $0.517$. Note that the p-value $=0.000812$ under the F-test for one-way ANOVA. Hence under $5\%$ level of significance, the average recall is significantly higher with Entropy. The improvement in recall can be explained by the fact that the keywords matches are guided so that the detections are guaranteed to have abundant keywords with variety of different species. Hence, the same noise level, the model with Entropy is able to capture more positive samples.

\renewcommand{\arraystretch}{1.25}
\begin{table}[h!]
\begin{tabular}{l c c} 
\Xhline{2\arrayrulewidth}
Category & Recall without Entropy & Recall with Entropy \\
\hline
business & 0.209 & \textbf{0.437}\\
tech & 0.023 & \textbf{0.406}\\
healthcare & 0.337 & \textbf{0.542}\\
legal & 0.239 & \textbf{0.460}\\
contract & 0.204 & \textbf{0.749}\\
hr & 0.146 & \textbf{0.374}\\
patent & 0.013 & \textbf{0.373}\\
tax & 0.430 & \textbf{0.838}\\
procurement & 0.344 & \textbf{0.470}\\
\Xhline{2\arrayrulewidth}
\end{tabular}
\caption{The table shows the recall of knowledge-based model with and without Shannon Entropy at a fixed FPR. With the Entropy, the recall of knowledge-based model is significantly increased over different target categories of content.}
\label{table1}
\end{table}
\renewcommand{\arraystretch}{1}

\renewcommand{\arraystretch}{1.25}
\begin{table}[h!]
\begin{tabular}{l c c c c} 
\Xhline{2\arrayrulewidth}
&  \multicolumn{2}{c}{Logistic Regression} & \multicolumn{2}{c}{Knowlegde-Based}\\
Category & {\footnotesize Recall on A}  & {\footnotesize Recall on B} & {\footnotesize Recall on A} & {\footnotesize Recall on B}\\
\hline
business & 0.615 & 0.273 & 0.526 & 0.318\\
tech & 0.941 & 0.280 & 0.635 & 0.272\\
healthcare & 0.482 & 0.049 & 0.898 & 0.367\\
legal & 0.291 & 0.126 & 0.577 & 0.350\\
contract & 0.780 & 0.579 & 0.761 & 0.749\\
hr & 0.572 & 0.176 & 0.551 & 0.283\\
patent & 0.448 & 0.055 & 0.378 & 0.378\\
tax & 0.899 & 0.556 & 0.840 & 0.856\\
procurement & 0.453 & 0.157 & 0.337 & 0.539\\
\Xhline{2\arrayrulewidth}
\end{tabular}
\caption{The table shows the change in recall for logistic regression and knowledge-based model with Shannon Entropy under change of data distribution from $A$ to $B$. While the performance of the logistic regression drastically decreases going from $A$ to $B$, the performance of knowledge-based model is comparatively robust under change of data distribution.}
\label{table2}
\end{table}
\renewcommand{\arraystretch}{1}

In the second experiment, we consider the effect of data variation on knowledge-based and machine learning models, particularly with limited positive samples for supervised training. To this end, we divide the document templates in each category into mutually disjoint template types $A$ and $B$, mimicking the change of data distribution from development to inferencing. We take the knowledge-based models with Entropy in the first experiment. Meanwhile, a logistic regression model is trained for each category based on the training split (half portion) of $A$ as positive samples and Wikipedia articles as negative samples. The prediction threshold is set by requiring FPR $=0.0005$ on the GPT-3 generated dataset.

Here, we study the robustness of the knowledge-based model, with Entropy with logistic regression as a benchmark. Recall of the knowledge-based model and logistic regression on validation split $A$ and all of $B$ are computed [Table \ref{table2}]. As seen from the results, the performance of the knowledge-based model is more robust over different categories while compared with the logistic regression. The average fractional change of recall from $A$ to $B$ is $-0.204$ and $-0.620$ for knowledge-based model and logistic regression respectively. The p-value $=0.0121$ for the F-test in one-way ANOVA, indicating a significant difference under $5\%$ significant level. While the data driven logistic model is fragile going from data distribution $A$ to $B$, the model performance is substantially more robust against change of data distribution for the knowledge-based model. This can be explained by the fact that the knowledge-based model does not learn on the data $A$. Instead, it is capturing the positive samples purely based on its exhaustive knowledge inside the model. Hence it has way better generalization power than data-driven approaches, particularly when the training dataset is small and not representative enough for machine learning.

\textbf{5.$~$Conclusion}---In this work, we present knowledge-based model with Shannon Entropy for document classification. The model intakes a glossary of domain specific keywords and scores the document content based on both abundance and diversity of the keywords. Such model provides a simple and explainable method for document classification. We show the model performance is notably better with Entropy. Moreover, the model is substantially more robust than the data-driven method.

The model here works well for broad content categories that can be identified by a rich set of specific keywords, but there are some limitations. First, some categories do not have many specific keywords. For example, press release content is generic and hence it is lacking of specific keywords. Second, some content are signalized by a small number of keyword matches. For example, offensive content is usually indicated by a single keyword of inappropriate language. Finally, the model is designed for capturing a broad scope of content. It is more difficult to precisely capture a granular category, where its glossary is usually a subset of keywords in a larger content category. For example, it is difficult to pull up information security policy documents without getting other types of information security documents. Hence, feasibility has to be checked in setting a target category.

While the main focus of this work is the implication of Entropy, there is room for improvement for the model. First, better abundance measures can be deployed. While the normalized tfidf score here is the most intuitive measure of concept abundance, there are more advanced measures, e.g. Okapi BM25 and its variations \cite{10.1561/1500000019}. Second, on top of the exact keyword matches here, semantic matches can also be used. It is well known that the best information retrieval solution is generally a combination of exact matches and semantic matches \cite{10.5555/1394399}. For example, embedding can be used for capturing the semantic representation of the text. Besides, modification can be made to enable granular document type classification. For example, setting keyword filters can lead to a more precise granular detection. These aspects will be future works along this line of research.

As a final remark, the application of Shannon Entropy is not restricted to document classification. Note that Entropy generally measures the information richness. Given a sample with a collection of positive features for recognition, it essentially measures the diversity of features showing up in it. In other words, Entropy itself can be a very informative feature for summarizing feature richness for artificial intelligence models and algorithms. While this work focuses on its application for recognition in text type of data, Entropy is expected to be applicable also for intelligence in image, time series and other generic type of data.

We thank fruitful interactions with Weisheng Li, Christian Rudnick, Michael Betser, Sharada Acharya, Sihong Liu, Ankit Srivastava from Microsoft Redmond team. We also thank Rajeethkumar Dharmaraj, Chithirai Meenal, Amit Kumar, Chinmaya Mishra from Microsoft IDC team for their thoughtful discussions.

\bibliography{ref}

\end{document}